\documentclass[runningheads]{llncs}
\usepackage{graphicx}

\usepackage[dvipsnames]{xcolor}

\usepackage{hyperref}
\hypersetup{colorlinks=true,
    linkcolor=black,
    filecolor=black,      
    urlcolor=black,
    citecolor=black}

\usepackage{amsmath}
\usepackage[sort]{cite}

\usepackage{tcolorbox}
\newcommand{\etc}{etc.\ }
\newcommand{\eg}{e.g., }

\newcommand{\ie}{i.e., }
\newcommand{\vs}{vs.\ }
\newcommand{\etal}{\textit{et al.}\ }

\usepackage[nameinlink]{cleveref}
\crefname{figure}{Fig.\hspace{-1pt}}{Figs.\hspace{-1pt}}
\Crefname{figure}{Figure}{Figures}
\crefname{equation}{Eq.\hspace{-1pt}}{Eqs.\hspace{-1pt}}
\Crefname{equation}{Equation}{Equations}
\crefname{section}{Section}{Sections}
\Crefname{section}{Section}{Sections}
\crefname{table}{Table}{Tables}
\crefname{appendix}{Appendix}{Appendices}

\newcommand{\chris}[1]{\textcolor{ForestGreen}{#1}}
\renewcommand{\chris}[1]{\textcolor[rgb]{0,0.0,0.0}{#1}}
\newcommand{\thms}[1]{\textcolor[rgb]{.8,0.3,0.1}{#1}}
\renewcommand{\thms}[1]{\textcolor[rgb]{0,0.0,0.0}{#1}}
\newcommand{\semere}[1]{\textcolor[rgb]{1,0.2,0.4}{#1}}
\renewcommand{\semere}[1]{\textcolor[rgb]{0,0.0,0.0}{#1}}

\newcommand{\semererev}[1]{\textcolor[rgb]{1,0.6,0.8}{#1}}
\renewcommand{\semererev}[1]{\textcolor[rgb]{0,0.0,0.0}{#1}}
\newcommand{\colortemplate}[1]{\textcolor{CadetBlue}{#1}}

\usepackage{booktabs}
\usepackage{amssymb}
\usepackage[inline,shortlabels]{enumitem}

\usepackage{multirow}
\usepackage{enumitem}
\usepackage{amsmath} 

\usepackage[colorinlistoftodos,prependcaption,textsize=normalsize]{todonotes}

\usepackage{todonotes}

\usepackage{graphicx}
\graphicspath{ {./figures/} }

\newcommand\dqsim{{\textsf{DQ-SIM}}}
\newcommand\mtfive{{\textsf{mT5}}}
\newcommand\zerochatgpt{{\textsf{Zero-ChatGPT}}}
\newcommand\staticchatgpt{{\textsf{Static-Demo-ChatGPT}}}
\newcommand\demochatgpt{{\textsf{Dynamic-Demo-ChatGPT}}}

\begin{document}
%

\title{Distractor generation for multiple-choice questions with predictive prompting and large language models }

%
\titlerunning{Distractor generation via predictive prompting and LLMs}
%

\author{Semere Kiros Bitew \and
Johannes Deleu \and
\\
Chris Develder \and
Thomas Demeester}

\authorrunning{Bitew et al.}
%
\institute{
IDLab, Ghent University -- imec, Ghent, Belgium\\
\email{\{semerekiros.bitew, johannes.deleu, chris.develder, thomas.demeester\}@ugent.be}}
\maketitle              
\begin{abstract}
   
Large Language Models (LLMs) such as ChatGPT have demonstrated remarkable performance across various tasks and have garnered significant attention from both researchers and practitioners.
\chris{However, in an educational context, we still observe a performance gap in generating distractors --- \ie plausible yet incorrect answers --- with LLMs for multiple-choice questions (MCQs).} In this study, we propose a strategy for guiding LLMs such as ChatGPT, in \chris{generating} relevant distractors by prompting them with question items automatically retrieved from a question bank as well-chosen in-context examples.
\chris{We evaluate our LLM-based solutions using a quantitative assessment on an existing test set, as well as through quality annotations by human experts, \ie teachers.}
\chris{We found that on average 53\% of the} 
\chris{generated} distractors presented to the teachers were rated as high-quality
\chris{, \ie} suitable for immediate use \chris{as is}, outperforming the state-of-the-art model. 
We also show the gains of our approach\footnote{{\url{https://github.com/semerekiros/distractGPT/} }} in generating high-quality distractors by comparing it with a zero-shot ChatGPT and a few-shot ChatGPT prompted with static examples. 

\end{abstract}
\section{Introduction}


The rapid advancement in artificial intelligence (AI) and large language models (LLMs) have paved the way for transformative applications across various domains, including the education domain. Since several \semere{LLMs (\eg GPT-3~\cite{brown2020language}, InstructGPT~\cite{ouyang2022training}, GPT-4~\cite{openai2023gpt4})} have been pretrained on massive amounts of data across multiple domains and languages, they are capable of solving natural language processing (NLP) tasks with little training examples (\ie few-shot learning) or no additional training (\ie zero-shot learning). This opens up new opportunities for adopting LLMs in the development of many educational technological solutions that aim to automate time-consuming and laborious educational tasks such as generating questions~\cite{kurdi2020systematic} and exercises
~\cite{bitew-etal-2023-learning}, essay scoring~\cite{ramesh2022automated}, and automated feedback~\cite{cavalcanti2021automatic}.  

In particular, the recent release of ChatGPT, an LLMs-based generative AI model that requires only natural language prompts without additional model training or fine-tuning, has demonstrated diverse potential in automating various educational tasks. For example, ChatGPT has achieved the equivalent of a passing score for a third-year medical student (above 60\%) in the United States Medical Licence Examination (USMLE) Step 1 exam, and provided logical justification and informational context across the majority of answers~\cite{gilson2023does}. Likewise, ChatGPT's performance on four real exams (containing 95 MCQs and 12 essay writing questions), at the University of Minnesota Law School was equivalent to C+ students implying a pass in the course~\cite{choi2023chatgpt}. \semere{Li~\etal~\cite{li2023can} show the capability} of ChatGPT in generating high-quality reflective responses in writing assignments administered for pharmacy courses.

One important educational task is the generation of multiple-choice questions (MCQs). MCQs have long been a popular form of formative and summative assessment in education due to their automatic scoring capability and the potential they hold for delivering timely and targeted feedback, which is crucial for facilitating effective learning~\cite{ramsden2003learning}. However, the process of crafting high-quality MCQs with effective distractors (\ie plausible yet incorrect answers) has traditionally been both a challenging and time-consuming task for educators (\eg teachers, content creators \etc) as poorly prepared distractors undermine the quality of MCQs~\cite{gierl2017developing}. This is where LLMs offer substantial benefits as they can be leveraged to automate the MCQ construction process, thus saving educators' time and effort while maintaining the quality and validity of the assessment items. For instance, teachers could employ LLMs to not only create different variants of the same MCQ questions but also develop different MCQs of comparable difficulty levels, facilitating targeted assessment for students with similar proficiency levels. Furthermore, students can benefit from the availability of several MCQs, enabling them to engage in regular practice, which is a well-established and highly effective learning strategy~\cite{roediger2006test}. Additionally, such models could be used for large-scale testing contexts (\eg licensure and certification testing) in which it is necessary to have multiple forms of a test and to introduce new question items regularly to minimize security concerns related to item exposure.    

In a recent study~\cite{bitew2022learning} conducted around the same time as the release of ChatGPT, researchers used local language models to automatically retrieve and reuse distractors to create new MCQs for education by leveraging existing pools of question items. In a user study they conducted with teachers, 3 out of 10 distractors proposed by their system were found to be high-quality, which is generally sufficient for creating an MCQ, as an average MCQ typically contains 3 distractors. However, they also report a staggering 50\% production of distractors that were entirely out of context given a question (so-called ``nonsense distractors''). With the emergence of ChatGPT, the question arises: \emph{does this previous approach become obsolete?} In our current study, we aim to address this question by examining the ability of out-of-the-box ChatGPT to generate effective distractors to be measured on the same scale as the previous study and evaluated by experts. Moreover, we study how both approaches could be combined into an even more effective approach. We also delve into the reliability issue, specifically in decreasing the production of nonsense distractors, which has implications for teachers' trust in the distractor generation tools.  
To guide our investigation, we formulate the following research questions (RQs):

\begin{enumerate}
    \item \label{item:rq1} \textbf{RQ1}: In comparison to ranking-based models, does ChatGPT generate high-quality distractors for educational MCQs?
    \item \label{item:rq2} \textbf{RQ2: }To what extent can we rely on ChatGPT-generated distractors, and how can we measure their trustworthiness? 
    \item \label{item:rq3} \textbf{RQ3: } Is it possible to enhance the capability of distractor generation by combining ranking-based models with LLMs?
\end{enumerate}

To answer the RQs, we designed ChatGPT prompting strategies and we solicited feedback from human experts, \ie teachers, to evaluate the quality of generated distractors. We also compared the different strategies in terms of the reliability of generating less nonsensical distractors. In general, we found ChatGPT-driven solutions produced high-quality distractors compared to ranking-based models. They are also more reliable than the ranking-based model as they produce significantly less number of nonsense distractors. 
\thms{We also combined the rank-based approach with ChatGPT, through the automatic composition of an example-based prompt from the output of the rank-based model.  We found that this}
leads to a more reliable and effective generation of distractors. The contribution of this paper can be summarized as follows:

\begin{itemize}

    \item We proposed a strategy to guide LLMs, specifically ChatGPT, to generate effective distractors for MCQs across various subjects by prompting the model with question items automatically retrieved from existing question banks. 
    \item We performed a user study with teachers to evaluate the quality of distractors proposed by our strategy.
    \item The evaluation of our approach unveils its dual capability to generate valuable distractors while simultaneously minimizing the occurrence of nonsensical options. 
\end{itemize}

The remainder of the paper is organized as follows: \cref{sec:related} describes the relevant work in distractor generation and LLM prompting strategies. \Cref{sec:method} explains the details of the baselines and the proposed method, while \cref{sec:experiments} introduces the test dataset and the evaluation setup of the user study with teachers. In \cref{sec:expertevaluation}, we report the results and provide some insights. Finally, in \cref{sec:conclusion}, we present the conclusion by summarizing the key findings and implications of our study.

\section{Related Work}
\label{sec:related}

Since we briefly covered the broad application of LLMs in education in the introduction, in this section we only focus on describing prior works on distractor generation (\cref{sec:distractorgeneration}) and discussing LLMs' prompting strategies (\cref{sec:promptingstrategies}) and their relevance to our work. 

\subsection{Distractor Generation}
\label{sec:distractorgeneration}

\chris{We focus} on generating incorrect options (\ie distractors) for multiple-choice questions (MCQs)
\chris{, which is} a time-consuming task that impacts MCQ quality and has been extensively researched. Broadly speaking, the main methods for generating distractors can be categorized into retrieval-based and generation-based techniques. 

\emph{Retrieval-based methods} generate distractors by selecting the most similar alternative answers in existing knowledge bases or question item corpora.
To approximate the similarity between distractors and the answer key (and question stem), several approaches are used 
based on
\begin{enumerate*}[(i)]
\item embedding space proximity~\cite{bitew2022learning,jiang2017distractor,guo2016questimator},
\item similarity in lexical databases such as WordNet~\cite{miller1995wordnet}, which is of particular importance in language and vocabulary learning~\cite{mitkov2009semantic,pino2008selection}, and
\item the semantic distance within domain-specific ontologies, which is critical in factoid-type questions~\cite{leo2019ontology,faizan2018automatic,alsubait2014generating,papasalouros2008automatic}.
\end{enumerate*}
This ultimately leads to the selection of candidate distractors based on a ranking strategy~\cite{liang2018distractor}.


\emph{Generation-based methods} make use of deep learning models to directly generate distractors.
Pioneering research~\cite{gao2019generating,yeung2019difficulty,zhou2020co} demonstrated the feasibility of using sequence-to-sequence models to generate distractors
\chris{, while more recently, solutions based on BERT~\cite{chung-etal-2020-bert,kalpakchi-boye-2021-bert} or T5~\cite{rodriguez2022end} have been explored.}
\chris{Rather than directly (auto-regressively) generate a distractor, the technique of back translation has shown to be relatively \thms{effective} (beating a BERT-based baseline) for fill-in-the-blank language assessment tests~\cite{panda-etal-2022-automatic}.}

In this work, we investigate the potential of ChatGPT\footnote{\url{https://chat.openai.com/}}, a large and autoregressive language model, in creating distractors.
We aim to combine retrieval-based and generative-based approaches by \begin{enumerate*}[(i)]
\item automatically retrieving similar question items from pre-existing question banks to compose an example prompt and \item using this example prompt to guide ChatGPT to generate relevant distractors.
    
\end{enumerate*}  







\subsection{Prompting strategies}
\label{sec:promptingstrategies}
\chris{Recent instruction-based large language models (LLMs) have been a game-changer for various tasks, showing remarkable performance without any task-specific training (\eg through finetuning) of the LLM~\cite{brown2020language,radford2019language}.
A specific task is solved through phrasing an instruction (zero-shot), possibly including a few input/output examples (few-shot) for the task at hand, as the so-called prompt that serves as input to the LLM.
The few-shot setting, including some examples, is commonly referred to as in-context learning (ICL).}
Another prompting strategy, chain-of-thought, induces language models to generate intermediate steps before predicting the final response~\cite{wei2022chain}.

In this paper, we introduce a variant of 
\chris{ICL} wherein the examples presented to the 
\chris{LLM} are 
determined \chris{dynamically,} based on the test example \chris{(\ie the question to generate distractors for, in our case)}.    


\section{Methods}
\label{sec:method}
We now describe our finetuned T5-based model (\cref{sec:t5-distractor-generation}), and out-of-the-box ChatGPT-based solutions in a zero-shot setting (\cref{sec:zero-shot-chatgpt}), as well as using in-context learning (\cref{sec:demonstration-chatgpt}).

\subsection{T5-based Distractor Generation}
\label{sec:t5-distractor-generation}

\chris{We} fine-tuned a multilingual T5 (mT5) model~\cite{xue-etal-2021-mt5} to generate distractors.
\chris{To this end, we use a private dataset (\ie the Televic dataset from ~\cite{bitew2022learning}) of 62K multiple-choice question items in the form of triplets comprising a question, answer and distractors.}
These question items are diverse in terms of language, domain, subject and question type. \semererev{On average, a question item has more than 2 distractors and contains exactly one answer. Additionally, the distractors in the dataset are not limited to single-word distractors. } 

\begin{figure}[h!]
\centering
\includegraphics[width=.6\textwidth]{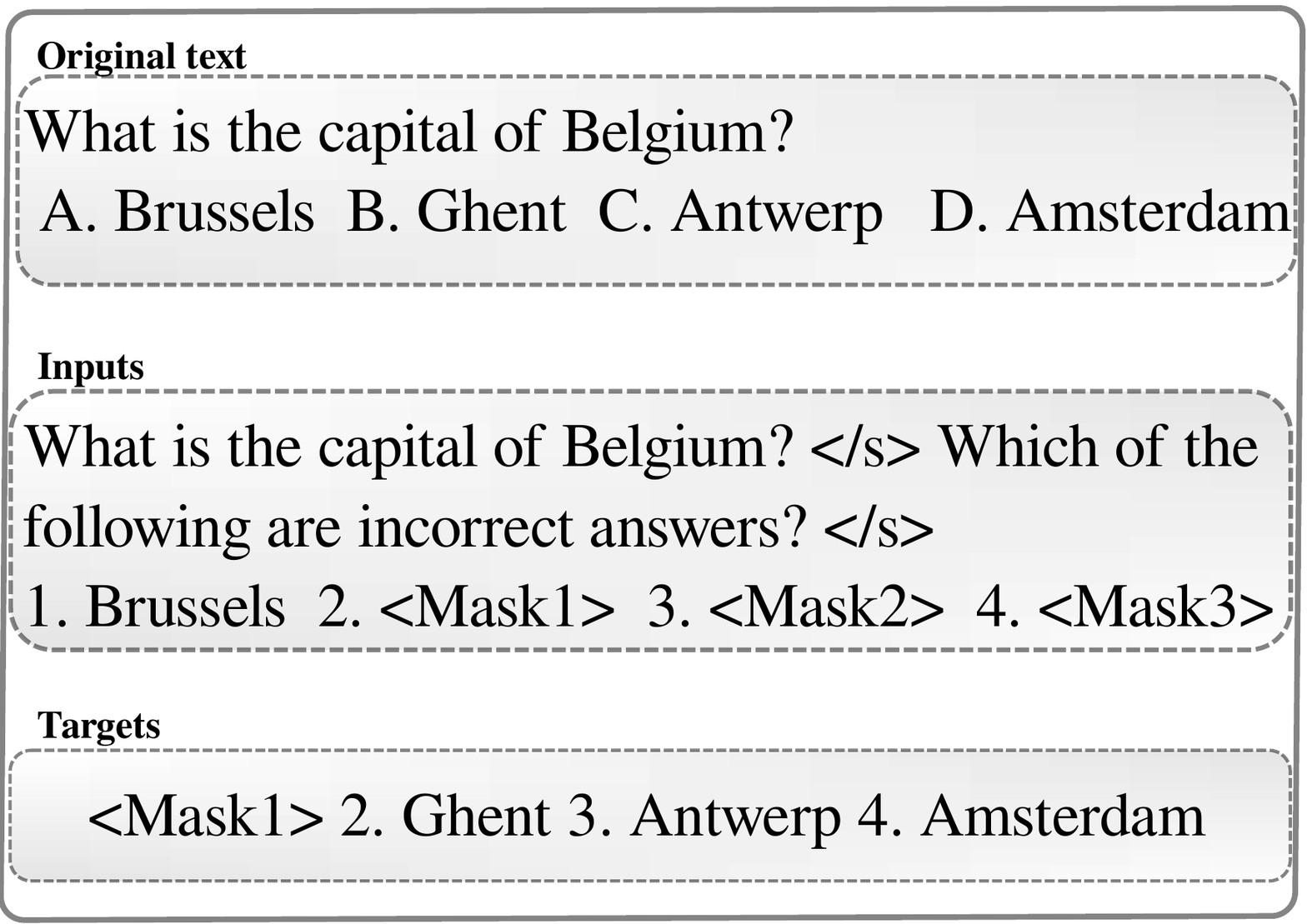}
\caption{Schematic of our fine-tuning procedure. The input sequence is constructed by copying the question and the answer from the original text and adding the template sentence \emph{``Which of the following are incorrect answers''}.
Each distractor is masked with a unique sentinel token (shown as 
\chris{$\langle$Mask$x\rangle$).}
The output sequence then consists of the dropped-out distractors.
Note that a single sentinel token replaces all consecutive spans of dropped-out tokens, and the template sentence is translated into the language of the question item (\ie Dutch or French). }
\label{fig:mt5template}
\end{figure}

Following the unsupervised pre-training objective used in the mT5 model, we rearranged our fine-tuning data into input and output sequences as illustrated in \cref{fig:mt5template}.
Our mT5 model's input sequence is constructed by copying the question stem and answer from the original question item and inserting the sentence \emph{``Which of the following are incorrect answers''} (or its translation depending on the language of the question item) between them.
Furthermore, we masked each distractor (\ie distractors could be multi-word spans) in the question item using a sentinel \chris{token}\footnote{Each sentinel token is assigned a token ID that is unique to the sequence.
The sentinel IDs are special tokens added to the model's vocabulary and do not correspond to any wordpiece} 
and separated them by 
\chris{increasing item numbers}.
The target sequence corresponds to all the dropped-out distractors and the objective is to predict the distractors.

The fine-tuning configuration that we have devised is intended to simplify the generation of multiple distractors. Specifically, all the necessary distractors for each question are generated as a list separated by numbers in a single decoding step.



\subsection{Zero-shot ChatGPT}
\label{sec:zero-shot-chatgpt}
\begin{figure}[h!]
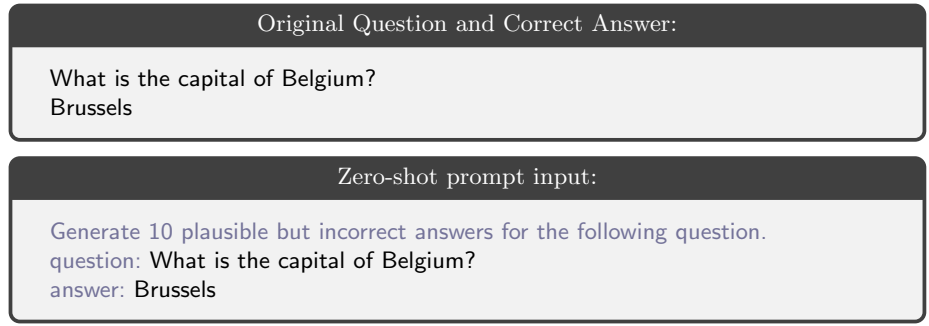

\centering
\begin{tcolorbox}[title=Original Question and Correct Answer:, center title]
\textsf{\small What is the capital of Belgium? \\  Brussels
}
\end{tcolorbox}
\begin{tcolorbox}[title=Zero-shot prompt input:, center title]
\textsf{\small \colortemplate{Generate 10 plausible but incorrect answers for the following question.}\\ 
\colortemplate{question:} What is the capital of Belgium? \\ 
\colortemplate{answer:} Brussels  
}
\end{tcolorbox}
\label{fig:zerotemplate}
\caption{Example of a question with its correct answer and how we turn that into a zero-shot prompt. 
Note that we translate the 
\colortemplate{fixed template parts} for questions in languages other than English.}
\label{fig:zero-shot-ChatGPT}
\end{figure}

\chris{To use ChatGPT in a zero-shot setting ({\zerochatgpt}),} we 
\chris{construct a prompt that concatenates} a fixed instruction sentence and the test example\chris{,} as shown in \chris{\cref{fig:zero-shot-ChatGPT}.} 
Note that each time a new query is made to ChatGPT, we clear conversations to avoid the influence of previous samples through independent API calls. We use a Python ChatGPT wrapper\footnote{Note that all the calls to the API were made between 06/04/2023 and 11/04/2023. Link to wrapper: \url{https://github.com/mmabrouk/chatgpt-wrapper}} to call the ChatGPT API automatically.  

\subsection{Demonstration-based ChatGPT }
\label{sec:demonstration-chatgpt}

\begin{figure}[t!]
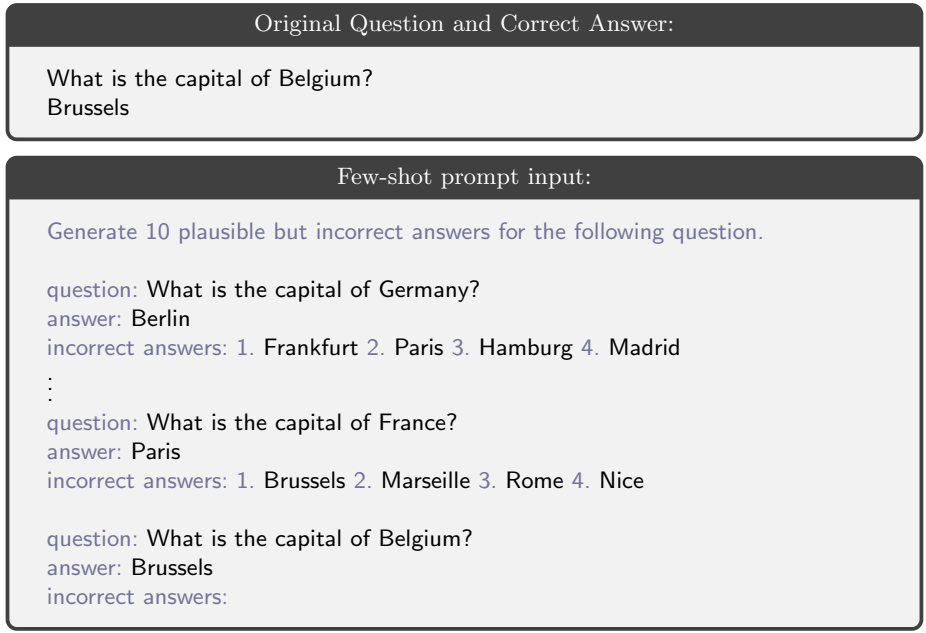

    \begin{tcolorbox}[title=Original Question and Correct Answer:, center title]
    \textsf{\small What is the capital of Belgium? \\  Brussels
    }
    \end{tcolorbox}
    \begin{tcolorbox}[title=Few-shot prompt input:, center title]
    \textsf{\small \colortemplate{Generate 10 plausible but incorrect answers for the following question.}\\
    \\
    \colortemplate{question:} What is the capital of Germany? \\ 
    \colortemplate{answer:} Berlin  \\
    \colortemplate{incorrect answers: 1.} Frankfurt 
    \colortemplate{2.} Paris
    \colortemplate{3.} Hamburg
    \colortemplate{4.} Madrid
    \\
    \vdots
    \\
    \colortemplate{question:} What is the capital of France? \\ 
    \colortemplate{answer:} Paris  \\
    \colortemplate{incorrect answers: 1.} Brussels 
    \colortemplate{2.} Marseille
    \colortemplate{3.} Rome
    \colortemplate{4.} Nice
    \\
    \\
    \colortemplate{question:} What is the capital of Belgium? \\ 
    \colortemplate{answer:} Brussels  \\
    \colortemplate{incorrect answers:}
    }
    \end{tcolorbox}
    \caption{Schematic of our demonstration-based prompt construction. The top-$k$ example demonstrations are automatically retrieved from the Televic question pool, and concatenated with the instruction and test instance.
    This prompt is used as a query to ChatGPT for generating distractors.
    Note that the 
    \chris{\colortemplate{fixed template} parts} are translated into the language of the test question item (\ie Dutch or French).}
    \label{fig:fewshot}
\end{figure}

Finally, we evaluate ChatGPT in a few-shot setting by probing it with smartly chosen demonstrations ({\demochatgpt}).
We propose to retrieve the most relevant question items from the Televic dataset (see \cref{sec:t5-distractor-generation}) and use them as demonstrations for a given test instance. We accomplish this by leveraging the question similarity (Q-SIM) model proposed by~\cite{bitew2022learning} to automatically select the top similar question items for the given test instance. \semererev{The Q-SIM model is a BERT-based ranking model that returns a ranked list of question items according to their similarity to a given test question.}
\chris{\Cref{fig:fewshot} illustrates how we combine the original question (to generate distractors for) with the retrieved examples into a prompt to ChatGPT.}

\section{Experiments}
\label{sec:experiments}

\subsection{Test Dataset}

\chris{To quantitatively evaluate our distractor generating models introduced in \cref{sec:method}, we}
use the Wezooz test data introduced by \cite{bitew2022learning}\chris{, which} 
comprises 300 multiple-choice questions (MCQs) designed for language and factual knowledge learning 
\chris{and is} aimed at secondary school students and teachers.
It includes French and English questions for language learning purposes, while Natural sciences, Geography, History and Biology constitute the factoid questions.
Each subject has 50 MCQs.
Note that the data distribution of the factoid questions is different from the Televic dataset (see \cref{sec:t5-distractor-generation} for details)
\chris{, which we use to}
\begin{enumerate*}[(i)]
\item fine-tune our mT5 model, and
\item 
\chris{retrieve} similar examples in our demonstration-based ChatGPT model.
\end{enumerate*}
However, the language learning questions are drawn from the same distribution, in a similar design setup as~\cite{bitew2022learning}.

\subsection{Human Expert Quality Assessment}
\label{sec:expert-eval-setup}
\chris{We also investigated our models' output quality using human assessors, by collecting feedback from teachers.}
For each 
\chris{of the 300 questions in the aforementioned WeZooz test set, we generated 10 distractors with each of our 3 models.}
The teachers were then presented with a randomized list of all 
\chris{30} generated distractors for each question.
They were explicitly instructed to rate each distractor independent of the other distractors in the list\chris{,} based on how much they thought it would help them if they were given the task of preparing distractors for that specific question. 
We used 
\chris{the} four-level annotation scheme proposed by \cite{bitew2022learning} to assign quality labels to each distractor\chris{:} 
\begin{enumerate*}[(1)]
    \item \textbf{True Answer}: the distractor partially or completely overlaps with the answer key.
    \item \textbf{Good distractor}: the distractor is viable and could be used in an MCQ as is.
    \item \textbf{Poor distractor}: the distractor is on topic but could easily be ruled out by students. 
    \item \textbf{Nonsense distractor}:  distractor is completely out of context.
\end{enumerate*}
\section{Results and Discussion}
\label{sec:resultsanddiscussion}

\begin{table}[t]
\small
\begin{center}
\caption{Inter-annotation agreement of experts
\chris{, measured by the} Jaccard similarity coeffic\chris{i}ent\chris{.}}
\label{tab:interannotation_jaccard}
\setlength{\tabcolsep}{6pt}
\begin{tabular}{lccccc}
\toprule
\textbf{Subjects} & 
\multicolumn{1}{c}{\textbf{True}} &
\multicolumn{1}{c}{\textbf{Good}} &
\multicolumn{1}{c}{\textbf{Poor}} &
\multicolumn{1}{c}{\textbf{Nonsense}} & 
\multicolumn{1}{c}{\textbf{Overall}} 
\\\midrule

English & 50.0 & 37.6 & 8.9 & 40.0 & 49.4\\
Geography & 33.3 & 75.7 & 34.1 & 35.0  & 74.0\\
\bottomrule
\end{tabular}
\end{center}

\end{table}

In this section, we provide evidence of the effectiveness and reliability of our approach by reporting the experimental results and discussing the insights obtained. In \cref{sec:interannotatoragreement}, we explain the annotation agreement among the teachers, followed by the evaluation results in \cref{sec:expertevaluation}. 

\subsection{Inter-annotator agreement}
\label{sec:interannotatoragreement}

Following the annotation scheme introduced in \cref{sec:expert-eval-setup}, a total of 12,860 ratings for distractor quality were collected from the annotation by teachers (see \cref{tab:ratings-data-description} in \cref{sec:appendix} for details of rating statistics). These ratings come from 10 distractors generated by each of the models (\ie all presented simultaneously to teachers as randomly shuffled list). \semererev{In total, 10 teachers participated in our quality assessment study.}

We adopt two strategies to determine the level of agreement between annotators. First, we ask teachers to rate the same set of distractors using the four-level annotation scale. We selected the subjects English, from language category, and Geography, from factoids, for annotations by at least two teachers. \Cref{tab:interannotation_jaccard} shows the inter-annotator agreement of teachers using the Jaccard similarity coefficient. The Jaccard similarity measures the similarity between two sets of data by calculating what fraction of the union of those datasets is covered by their intersection. In our case, it is calculated as the number of times the teachers agreed on a distractor quality label (\ie one of the four labels), divided by the total number of distractors that were annotated (by either annotator) with that label. \semere{In general, we note a higher agreement on what is considered a good distractor compared to the other distractor categories. Moreover, the overall agreement between the Geography teachers is higher than the English teachers.} 

\semere{Second, we employed the widely utilized Cohen's kappa coefficient \cite{mchugh2012interrater}. Our analysis substantiates the previously mentioned observation that annotators have a greater consensus when evaluating factoid questions compared to language-related queries as \cite{bitew2022learning}. Specifically, among English teachers, the calculated Cohen's kappa value stands at 28.9, signifying a ``fair agreement'' level. Similarly, Geography teachers exhibit a higher level of agreement with a Cohen's kappa value of 52, indicating a level of agreement categorized as ``moderate.'' 
}

\subsection{Evaluation of models}
\label{sec:expertevaluation}

\begin{table*}[t]
\centering
\caption{Expert evaluation of distractors (\%). GDR: good distractor rate, NDR: nonsense distractor rate; $\uparrow$: higher is better, $\downarrow$: lower is better; evaluation on WeZooz test set. The markers $\star$ and $\ddag$ respectively denote the one-tailed significance levels of the bootstrap-based $p-$value, \ie $p<$ 0.1 and $p<$ 0.01 with respect to the best model {\demochatgpt} in each column. }
\label{tab:results_expert_eval}
\setlength{\tabcolsep}{4pt}
\small
\begin{tabular}{lcccc}
\toprule
\textbf{Models} & 
\multicolumn{2}{c}{\textbf{Language learning}} &
\multicolumn{2}{c}{\textbf{Factoid learning }} 
\\
\cmidrule(lr){2-3} \cmidrule(lr){4-5}
& \multicolumn{1}{c}{\textbf{GDR@10\,$\uparrow$}} &
\multicolumn{1}{c}{\textbf{NDR@10\,$\downarrow$}} &
\multicolumn{1}{c}{\textbf{GDR@10\,$\uparrow$}} &
\multicolumn{1}{c}{\textbf{NDR@10\,$\downarrow$}} 
\\
\midrule
{\dqsim} \cite{bitew2022learning} & 27.9$^\ddag$ & 44.6$^\ddag$ & 28.9$^\ddag$ & 50.1$^\ddag$   \\ 
{\mtfive} & 24.5$^\ddag$ & 42.3$^\ddag$ & 27.8$^\ddag$ & 36.6$^\ddag$ \\ 
{\zerochatgpt} & 30.2$\ddag$ & 34.6$\ddag$ & 57.6$\star$ & 17.5$\star$ \\ 
{\demochatgpt} & \textbf{46.7} & \textbf{15.5} &  \textbf{58.8} & \textbf{16.4} \\ 
\bottomrule
\end{tabular}

\end{table*}

\Cref{tab:results_expert_eval} shows the expert evaluation of distractors in terms of \emph{good distractor rate} (GDR@10), and \emph{nonsense distractor rate} (NDR@10). GDR@10 is calculated as the percentage of distractors that were rated `good' among the proposed 10 distractor for each model. Similarly, NDR@10 is calculated as the percentage of distractors that were rated `nonsense' among the 10 candidate distractors proposed by each model. We are interested in reporting the NDR metric 
because it could be used as a measure of the reliability of educational models, as a high occurrence of nonsense distractors may undermine users' trust in the model. 
The reported metrics are averages of all the subjects in each category (\ie French and English for language learning, and Biology, Natural Sciences, History and Geography for factoids). In the table, the upward arrow ($\uparrow$) indicates larger values are desired, while the downward arrow$\downarrow$ indicates smaller values are preferred. 

In general, the ChatGPT-based solutions (\ie {\zerochatgpt}, and {\demochatgpt}) were rated better in proposing plausible distractors than the baselines. They also produced fewer nonsense distractors. Particularly, the {\demochatgpt} outperformed all the other models. On average, approximately 5 of its 10 proposed distractors were rated high-quality distractors and only 1.5 distractors were rated nonsense. Moreover, on average 8.5 distractors were generally found to be on-topic (\ie distractors rated as either good or poor distractors) for our best model {\demochatgpt}.

All the models are better at generating effective distractors for factoids than for language questions as shown by the higher GDR@10 results for factoids than languages. \semere{We hypothesize this is because, for factoid questions, our models are mainly tasked with generating accurately composed distractors that are contextually incorrect. In contrast, when faced with language questions, the intended distractors may possess ungrammatical attributes, posing a challenge for our models to generate text that is intentionally ungrammatical.}



Our purely generative local {\mtfive} model does not improve the {\dqsim} model (\ie previous state-of-the-art model on the test set) at proposing good distractors (\ie GDR@10 of 24.5 \vs 27.9 and 28.9 \vs 27.8). However, it is a more reliable model as it produces fewer nonsense distractors as illustrated by its lower NDR@10 values of 42.3 and 36.6 for languages and factoids, respectively, in contrast to the corresponding values of 44.6 and 50.1 for the {\dqsim} model. The relatively high number of nonsense distractors in {\dqsim} is partly attributed to its inherent limitation of only ranking pre-existing distractors according to their relevance to a given question, thereby lacking the ability to generate brand-new distractors.


\semere{In addition, in order to ensure the validity of the differences between the models, we carry out a bootstrap significance analysis~\cite{sakai2007evaluating} by sampling with replacement the annotation results {\dqsim}, {\mtfive}, {\zerochatgpt}, and {\demochatgpt} models 1000 times. The resulting one-tailed significance levels ($p$ values) are indicated in \cref{tab:results_expert_eval} by markers $\star$ and $\ddag$ which respectively denote $p<0.1$ and $p<0.01$ with respect to our best model {\demochatgpt{}} in each column.}


\paragraph{\textbf{Effect of dynamically retrieved in-context examples}} We replace the dynamically retrieved examples with randomly selected \semere{language} in-context examples from the Televic question bank, and we keep this selection constant (\ie {\staticchatgpt}) to generate distractors. Similar to the other models, we generated 10 distractors using the {\staticchatgpt} model and asked teachers to annotate the quality of the distractors. We focused on the language learning category as it showed a huge performance improvement when transitioning from {\zerochatgpt} to {\demochatgpt}.  

We observe that the {\demochatgpt{}} model significantly outperforms the {\staticchatgpt{}} model in generating high-quality distractors as indicated by the GDR@10 metric in \cref{tab:ablation}. However, the difference in generating less nonsense distractor (\ie NDR@10) is not significant. See \cref{tb:generated_examples} in \cref{sec:appendix} for an example of generated distractors using the approaches.  
\begin{table}[t]
\small
\begin{center}
\caption{Effect of using dynamically retrieved in-context examples: \demochatgpt{} \vs \staticchatgpt{} that uses static in-context examples for language learning. The markers $\ddag$ denotes the one-tailed significance level of the bootstrap-based $p-$value, \ie $p<0.01$ with respect to {\demochatgpt}}
\label{tab:ablation}
\setlength{\tabcolsep}{6pt}
\begin{tabular}{lcc}
\toprule
\textbf{Models} & 
\textbf{GDR@10\,$\uparrow$} &
\textbf{NDR@10\,$\downarrow$} 
\\\midrule
{\staticchatgpt} & 43.3$\ddag$ & 16.2  \\ 
{\demochatgpt} & \textbf{46.7} & 15.5 \\ 
\bottomrule
\end{tabular}
\end{center}

\end{table}

\subsection{Discussion of Research questions}
To answer \textbf{RQ\ref{item:rq1}}, we compare the ChatGPT-based solutions (\ie {\zerochatgpt}, {\staticchatgpt{}} and {\demochatgpt{}}) with the previous state-of-the-art ranking-based model, {\dqsim{}} in generating distractors. All the ChatGPT-based distractor generation strategies significantly outperform the {\dqsim}.

To address \textbf{RQ\ref{item:rq2}}, we employ the NDR@10 metric as a proxy to measure the trustworthiness of models. Our best model produces an average of only 16\% nonsense distractors, which is a remarkable improvement compared to the previously reported state-of-the-art performance of 50\% NDR@10. This significant reduction of nonsense distractors can be expected to inspire more trust in the approach by teachers. 

To answer \textbf{RQ\ref{item:rq3}}, we compare {\demochatgpt{}}, which combines a local ranking model with ChatGPT, against {\zerochatgpt} and {\staticchatgpt{}}. As shown in \cref{tab:results_expert_eval} and \cref{tab:ablation}, combining local models with ChatGPT leads to a better quality distractor generation, highlighting the effectiveness of this combined approach. 

\section{Conclusion}
\label{sec:conclusion}

This research paper introduced and evaluated a novel strategy designed to guide LLMs, such as ChatGPT, in generating reliable and effective distractors for the creation of MCQs in educational contexts. Our proposed approach, \demochatgpt{} model combines a rank-based approach with ChatGPT. This involves the dynamic retrieval of relevant question items through the ranker that are then presented as in-context examples to ChatGPT for generating distractors. Our results indicated that the \demochatgpt{} showed a considerably reduced production of nonsense distractors (\ie only 16\% rated as nonsense) compared to \zerochatgpt{} (\ie out-of-the-box ChatGPT), which we consider a useful asset in terms of trust in the model by teachers. Moreover, on average, 5 out of the 10 distractors suggested by our approach were rated as high-quality by teachers, to be readily used.

\semererev{For future work, we aim to investigate designing a fine-grained evaluation setup for distractors that takes into account various factors such as the level of the student, the difficulty of the questions etc. There is also a potential to explore alternative prompting strategies for LLMS, when generating distractors. For example, the utilization of self-correcting mechanism~\cite{wang2023self}, which involves revising the initial output of an LLM by evaluating certain aspects of the text, could be explored in the context of distractor generation.}

\bibliographystyle{splncs04}
\bibliography{mybibliography}
\appendix
\section{User Study Details}
\label{sec:appendix}

This section contains the user study details. \Cref{tab:ratings-data-description} describes the data gathered from the annotations provided by teachers. Every subject contains 50 questions, except English which has 48 questions. We collected 12,860 annotations for the proposed candidate distractors (\ie 10 distractors by each of the three models). A total of 10 teachers participated in the study. English (\ie from languages) and Geography (\ie from factoids) were annotated twice by two different teachers to calculate inter-annotator agreement. \semere{Additionally, to study the effect of dynamic retrieval of in-context examples, we asked 1 English and 1 French teacher to annotate the distractor predictions from the {\staticchatgpt{}} model.} \semererev{The second column (\ie \emph{Item count}), shows the number of question items for each subject in the Wezooz dataset. Alongside, the \emph{distractors count} column provides two distinct values: the gold truth distractors count within the dataset, and the count of unique distractors generated by our models. It is important to note that different models may produce identical distractors for a given question, resulting in varying numbers of newly generated distractors across the different subjects.} 


\begin{table}[ht!]
\centering
\footnotesize
\begin{center}
\caption{Ratings Data Description }
\label{tab:ratings-data-description}

\setlength{\tabcolsep}{5pt}

\begin{tabular}{lccccccc}

\toprule
\textbf{Subjects} & 
\multicolumn{1}{c}{\textbf{Item count}} &
\multicolumn{2}{c}{\textbf{Distractors count }} &
\multicolumn{1}{c}{\textbf{Ratings count}} &
\multicolumn{1}{c}{\textbf{No of Raters}} &
\\\cmidrule(lr){3-4} 
& &
\multicolumn{1}{c}{\textbf{Gold}} &
\multicolumn{1}{c}{\textbf{Generated}} &

\\\midrule
English         &  48   & 130   & 1324      & 3360  & 3 \\ 
French          &  50   & 92    & 1396      & 2000  & 2 \\ 
Geography       &  50   & 145   & 1248      & 3000  & 2 \\ 
History         &  50   & 130   & 1354     & 1500  & 1 \\ 
Biology         &  50   & 88    & 1391      & 1500  & 1 \\ 
Nat. Sciences   & 50    & 100   & 1449     & 1500  & 1 \\
\midrule
Total           & 298   & 685   & 8162  & 12860   & 10  \\ 
\bottomrule
\end{tabular}
\end{center}

\end{table}

\section{Example Generated Distractors}
\label{sec:generatedexamples}

\semere{\Cref{tb:generated_examples} shows a cherry-picked example of our ChatGPT-based models. }

\begin{table*}[t]

    \caption{Some generated examples from {\zerochatgpt}, {\demochatgpt}, {\staticchatgpt} models for English. High-quality distractors are shown in \textbf{boldface}, while on-topic and nonsense distractors are \textit{italicized} and \underline{underlined}, respectively. We only show 2 in-context examples for the \staticchatgpt and \demochatgpt models as part of the prompt but in practice, we use 5 of such examples.}
    \label{tb:generated_examples}
    \centering
    \setlength{\tabcolsep}{5pt}

    \begin{tabular}{p{0.1\linewidth} p{0.8\linewidth}}
\toprule

\multicolumn{1}{l}{\textbf{Model}} &  
\multicolumn{1}{c}{\textbf{Example}} \\

\midrule
\zerochatgpt & \textbf{Prompt}: Generate 10 plausible but incorrect answers for the following question. \newline
question: They do n't speak English and can not make ... understood. \newline 
answer: themselves      \newline
      \textbf{Ground}: [them, us,you]     \newline
      \textbf{Prediction}: [1. \emph{each other}, 2. \underline{any sense}, 3. \underline{a sentence}, 4. \underline{appropriate gestures}, 5. \underline{complete thoughts}, 6. \underline{proper communication}, 7. \underline{accurate translations}, 8. \underline{meaningful conversation}, 9. \underline{clear statements}, 10. \underline{intelligible sounds}]
       \\
\midrule
\staticchatgpt &\textbf{Prompt}: Generate 10 plausible but incorrect answers for the following question. \newline
question: Why do n't we ask Jane to come over and talk ? She 's an ... \newline
answer:  alumna \newline 
incorrect answers: 1. alumnus  2. alumni  3. alumn \newline 
question: Choose the opposite of the word in bold . This company is loss-making . \newline
answer:  profit-making \newline 
incorrect answers: 1. loose  2. lucrative  3. unprofitable \newline 

. . . \newline

question: They do n't speak English and can not make ... understood. \newline 
answer: themselves      \newline
      \textbf{Ground}: [them, us,you]     \newline
      \textbf{Prediction}: [1. \emph{themselvess}, 2. \textbf{themselves} 3. \textbf{themselfs}, 4. \textbf{themselfes}, 5. \emph{themselfis}, 6. \emph{themselfies}, 7. \textbf{themselfves}, 8. \emph{themselfselves}, 9. \underline{themselfiness}, 10. \underline{themselfulated}
]
      \\
\midrule
\demochatgpt &\textbf{Prompt}: Generate 10 plausible but incorrect answers for the following question. \newline
question: They should choose ... would be best suited for the job \newline
answer:  which \newline 
incorrect answers: 1. whom  2. that  3. him who \newline 
question: 36 . He does not speak English and can not make ... understood .\newline
answer:  himself \newline 
incorrect answers: 1. you  2. us  3. herself \newline 
question: The new e-ticketing system , ... ... ... ... was installed last month , is working well . \newline
answer:  which \newline 
incorrect answers: 1. what  2. him  3. where \newline

... \newline

question: They do n't speak English and can not make ... understood. \newline 
answer: themselves      \newline
      \textbf{Ground}: [them, us,you]     \newline
      \textbf{Prediction}: [1.\textbf{theirself}, 
2. \textbf{theirselves},
3. \textbf{themself},
4. \textbf{themselfs}, 
5. \textbf{theirs}, 
6. \textbf{us},
7. \underline{they}, 
8. \underline{them is}, 
9. \textbf{theyself}, 
10. \textbf{themselfes}
]
      \\
    \bottomrule

    \end{tabular}
\end{table*}

\end{document}